\documentclass{intech07}

\usepackage{amsmath}
\usepackage{rotating}
\usepackage{array}
\usepackage{booktabs}
\usepackage{multirow}
\newcommand{\alsp}{ALIZE/SpkDet~}
\newcommand{\bx}{\boldsymbol x}

\booktitle{Biometrics / Book 1}%

\chaptertitle{Human Identity Verification based on Heart Sounds: Recent Advances
and Future Directions} 

\authors{Francesco Beritelli and Andrea Spadaccini}
\affiliation{Dipartimento di Ingegneria Elettrica, Elettronica ed Informatica
(DIEEI)\\University of Catania}
\country{Italy}

\begin{document}
\maketitle

\section{Introduction}
\label{sec:intro}
Identity verification is an increasingly important process in our daily lives.
Whether we need to use our own equipment or to prove our identity to third
parties in order to use services or gain access to physical places, we are
constantly required to declare our identity and prove our claim.

Traditional authentication methods fall into two categories:
proving that you know something (i.e., password-based authentication)
and proving that you own something (i.e., token-based authentication).

These methods connect the identity with an alternate and less rich
representation, for instance a password, that can be lost, stolen, or shared.

A solution to these problems comes from biometric recognition systems.
Biometrics offers a natural solution to the authentication problem, as it
contributes to the construction of systems that can recognize people by the
analysis of their anatomical and/or behavioral characteristics. With biometric
systems, the representation of the identity is something that is directly
derived from the subject, therefore it has properties that a surrogate
representation, like a password or a token, simply cannot have
(\cite{prabhakar-privacy,intro-biometrics,biometrics-security}).

The strength of a biometric system is determined mainly by the trait that is
used to verify the identity. Plenty of biometric traits have been studied and some
of them, like fingerprint, iris and face, are nowadays used in widely deployed
systems.

Today, one of the most important research directions in the field of
biometrics is the characterization of novel biometric traits that can be used
in conjunction with other traits, to limit their shortcomings or to enhance
their performance.

The aim of this chapter is to introduce the reader to the usage of heart
sounds for biometric recognition, describing the strengths and the
weaknesses of this novel trait and analyzing in detail the methods developed
so far and their performance.

The usage of heart sounds as physiological biometric traits was first
introduced in \cite{heart-bio}, in which the authors proposed and started
exploring this idea. Their system is based on the frequency analysis, by
means of the Chirp $z$-Transform (CZT), of the sounds produced by the heart
during the closure of the mitral tricuspid valve and during the closure of the
aortic pulmonary valve. These sounds, called S1 and S2, are extracted
from the input signal using a segmentation algorithm. The authors build the identity
templates using feature vectors and test if the identity claim is true by
computing the Euclidean distance between the stored template and the features
extracted during the identity verification phase.

In \cite{heart-singapore}, the authors describe a different approach to
heart-sounds biometry. Instead of doing a structural analysis of the input
signal, they use the whole sequences, feeding them to two recognizers built
using Vector Quantization and Gaussian Mixture Models; the latter proves to be
the most performant system.

In \cite{heart-dsp2009,heart-wifs09}, the authors further develop the
system described in \cite{heart-bio}, evaluating its performance on a
larger database, choosing a more suitable feature set (Linear Frequency
Cepstrum Coefficients, LFCC), adding a time-domain feature specific for heart
sounds, called First-to-Second Ratio (FSR) and adding a quality-based data
selection algorithm.

In \cite{heart-securware2010,heart-bioms2010}, the authors take an alternative
approach to the problem, building a system that leverages statistical modelling
using Gaussian Mixture Models. This technique is different from
\cite{heart-singapore} in many ways, most notably the segmentation of the
heart sounds, the database, the usage of features specific to heart sounds and
the statistical engine. This system proved to yield good performance in spite
of a larger database, and the final Equal Error Rate (EER) obtained using this
technique is 13.70 \% over a database of 165 people, containing two heart
sequences per person, each lasting from 20 to 70 seconds.

This chapter is structured as follows: in Section~\ref{sec:heart-sounds-biometry},
we describe in detail the usage of heart sounds for biometric identification,
comparing them to other biometric traits, briefly explaining
how the human cardio-circulatory system works and produces
heart sounds and how they can be processed; in Section~\ref{sec:review} we
present a survey of recent works on heart-sounds biometry by other research
groups; in Section~\ref{sec:structural-approach} we describe in detail the
structural approach; in Section~\ref{sec:statistical-approach} we describe the
statistical approach; in Section~\ref{sec:comparison} we compare the
performance of the two methods on a common database, describing both the
performance metrics and the heart sounds database used for the
evaluation; finally, in Section~\ref{sec:conclusions} we present our
conclusions, and highlight current issues of this method and suggest the
directions for the future research.

\section{Biometric recognition using heart sounds}
\label{sec:heart-sounds-biometry}
Biometric recognition is the process of inferring the identity of a person via
quantitative analysis of one or more traits, that can be derived either
directly from a person's body (physiological traits) or from one's behaviour
(behavioural traits).

Speaking of physiological traits, almost all the parts of the body can already
be used for the identification process (\cite{handbook-biometrics}): eyes (iris
and retina), face, hand (shape, veins, palmprint, fingerprints), ears, teeth
etc.

In this chapter, we will focus on an organ that is of fundamental importance
for our life: the heart.

The heart is involved in the production of two biological signals, the
Electrocardiograph (ECG) and the Phonocardiogram (PCG). The first is a signal
derived from the electrical activity of the organ, while the latter is a
recording of the sounds that are produced during its activity (heart sounds).

While both signals have been used as biometric traits (see \cite{biel-ecg} for ECG-based
biometry), this chapter will focus on hearts-sounds biometry.

\subsection{Comparison to other biometric traits}
\label{sec:traits-comparison}
The paper~\cite{intro-biometrics} presents a classification of available
biometric traits with respect to 7 qualities that, according to the authors, a
trait should possess:
\begin{itemize}
\vspace{.5mm}    \item{\textbf{Universality}: each person should possess it;}
    \item{\textbf{Distinctiveness}: it should be helpful in the
        distinction between any two people;}\vspace{.5mm}
    \item{\textbf{Permanence}: it should not change over time;}\vspace{1mm}
    \item{\textbf{Collectability}: it should be quantitatively measurable;}\vspace{.5mm}
    \item{\textbf{Performance}: biometric systems that use it should be\vspace{.5mm}
        reasonably performant, with respect to speed, accuracy and computational
        requirements;}\vspace{.5mm}
    \item{\textbf{Acceptability}: the users of the biometric system should see
        the usage of the trait as a natural and trustable thing to do in order to
        authenticate;}\vspace{.5mm}
    \item{\textbf{Circumvention}: the system should be robust to malicious
        identification attempts.}\vspace{.5mm}
\end{itemize}
Each trait is evaluated with respect to each of these qualities using 3 possible
qualifiers: H (high), M (medium), L (low).

We added to the original table a row with our subjective evaluation of heart-sounds
biometry with respect to the qualities described above, in order to compare
this new technique with other more established traits. The updated table is reproduced in
Table~\ref{table:traits_comparison}.

The reasoning behind each of our subjective evaluations of the qualities of
heart sounds is as follows:

\begin{itemize}
    \item{\textbf{High Universality}: a working heart is a \textit{conditio
        sine qua non} for human life;}\vspace{.5mm}
    \item{\textbf{Medium Distinctiveness}: the actual systems' performance is
        still far from the most discriminating traits, and the tests are
        conducted using small databases; the discriminative power of heart sounds
        still must be demonstrated;}\vspace{.5mm}
    \item{\textbf{Low Permanence}: although to the best of our knowledge
        no studies have been conducted in this field, we perceive that heart sounds
        can change their properties over time, so their accuracy over extended time
        spans must be evaluated;}\vspace{.5mm}
    \item{\textbf{Low Collectability}: the collection of heart sounds is not an
        immediate process, and electronic stethoscopes must be placed in well-defined
        positions on the chest to get a high-quality signal;}\vspace{.5mm}
    \item{\textbf{Low Performance}: most of the techniques used for
        heart-sounds biometry are computationally intensive and, as said before, the
        accuracy still needs to be improved;}\vspace{.5mm}
    \item{\textbf{Medium Acceptability}: heart sounds are probably identified
        as unique and trustable by people, but they might be unwilling to use them
        in daily authentication tasks;}\vspace{.5mm}
    \item{\textbf{Low Circumvention}: it is very difficult to reproduce the
        heart sound of another person, and it is also difficult to record it covertly
        in order to reproduce it later.}
\end{itemize}

\begin{table}[ht]
    \centering
    \begin{tabular}{l|>{\centering\arraybackslash}m{0.35in}|>{\centering\arraybackslash}m{0.35in}|>{\centering\arraybackslash}m{0.35in}|>{\centering\arraybackslash}m{0.35in}|>{\centering\arraybackslash}m{0.35in}|>{\centering\arraybackslash}m{0.35in}|>{\centering\arraybackslash}m{0.35in}}
        \hline\hline
        Biometric identifier &
        \begin{sideways}Universality\end{sideways} &
        \begin{sideways}Distinctiveness\,\end{sideways} &
        \begin{sideways}Permanence\end{sideways} &
        \begin{sideways}Collectability\end{sideways} &
        \begin{sideways}Performance\end{sideways} &
        \begin{sideways}Acceptability\end{sideways} &
        \begin{sideways}Circumvention \end{sideways}\\
        \hline
        DNA                & H & H & H & L & H & L & L\\
        Ear                & M & M & H & M & M & H & M\\
        Face               & H & L & M & H & L & H & H\\
        Facial thermogram  & H & H & L & H & M & H & L\\
        Fingerprint        & M & H & H & M & H & M & M\\
        Gait               & M & L & L & H & L & H & M\\
        Hand geometry      & M & M & M & H & M & M & M\\
        Hand vein          & M & M & M & M & M & M & L\\
        Iris               & H & H & H & M & H & L & L\\
        Keystroke          & L & L & L & M & L & M & M\\
        Odor               & H & H & H & L & L & M & L\\
        Palmprint          & M & H & H & M & H & M & M\\
        Retina             & H & H & M & L & H & L & L\\
        Signature          & L & L & L & H & L & H & H\\
        Voice              & M & L & M & L & L & M & H\\
        \hline
        Heart sounds       & H & M & L & L & L & M & L\\
        \hline
    \end{tabular}
        \caption{Comparison between biometric traits as in \cite{intro-biometrics} and heart sounds}    \label{table:traits_comparison}
\end{table}

Of course, heart-sounds biometry is a new technique, and some of its
drawbacks probably will be addressed and resolved in future research work.

\subsection{Physiology and structure of heart sounds}
\label{sec:heart-sounds-physiology}
The heart sound signal is a complex, non-stationary and quasi-periodic signal
that is produced by the heart during its continuous pumping work
(\cite{robust-hsad-2010}). It is composed by several smaller sounds, each
associated with a specific event in the working cycle of the heart.

Heart sounds fall in two categories:

\begin{itemize}
    \item{\textbf{primary sounds}, produced by the closure of the heart valves;}
    \item{\textbf{other sounds}, produced by the blood flowing in the heart or
by pathologies;}
\end{itemize}

The primary sounds are S1 and S2. The first sound, S1, is caused by the closure
of the tricuspid and mitral valves, while the second sound, S2, is caused by
the closure of the aortic and pulmonary valves.

Among the other sounds, there are the S3 and S4 sounds, that are quieter and
rarer than S1 and S2, and murmurs, that are high-frequency noises.

In our systems, we only use the primary sounds because they
are the two loudest sounds and they are the only ones that a heart
always produces, even in pathological conditions. We separate them from the rest
of the heart sound signal using the algorithm described in Section~\ref{sec:segmentation}.

\subsection{Processing heart sounds}
\label{sec:heart-sounds-processing}
Heart sounds are monodimensional signals, and can be processed, to some
extent, with techniques known to work on other monodimensional signals, like
audio signals. Those techniques then need to be refined taking into account the
peculiarities of the signal, its structure and components.

In this section we will describe an algorithm used to separate the S1 and S2
sounds from the rest of the heart sound signal (\ref{sec:segmentation}) and
three algorithms used for feature extraction (\ref{sec:czt},
\ref{sec:cepstral}, \ref{sec:fsr}), that is the process of transforming the
original heart sound signal into a more compact, and possibly more meaningful,
representation. We will briefly discuss two algorithms that work in the
frequency domain, and one in the time domain.

\subsubsection{Segmentation}
\label{sec:segmentation}
In this section we describe a variation of the algorithm that was employed in
(\cite{heart-bio}) to separate the S1 and S2 tones from the rest of the heart
sound signal, improved to deal with long heart sounds.

Such a separation is done because we believe that the S1 and S2 tones are
as important to heart sounds as the vowels are to the voice signal. They are stationary in
the short term and they convey significant biometric information, that is then
processed by feature extraction algorithms.

A simple energy-based approach can not be used because the signal can
contain impulsive noise that could be mistaken for a significant
sound.

The first step of the algorithm is searching the frame with the highest
energy, that is called SX1. At this stage, we do not know if we found an S1 or
an S2 sound.

Then, in order to estimate the frequency of the heart beat, and therefore the
period $P$ of the signal, the maximum value of the autocorrelation function is
computed. Low-frequency components are ignored by searching
only over the portion of autocorrelation after the first minimum.

The algorithm then searches other maxima to the left and to the right of SX1,
moving by a number $P$ of frames in each direction and searching for local
maxima in a window of the energy signal in order to take into account small
fluctuations of the heart rate. After each maximum is selected, a
constant-width window is applied to select a portion of the signal.

After having completed the search that starts from SX1, all the corresponding
frames in the original signal are zeroed out, and the procedure is repeated to
find a new maximum-energy frame, called SX2, and the other peaks are found in
the same way.

Finally, the positions of SX1 and SX2 are compared, and the algorithm then
decides if SX1, and all the frames found starting from it, must be classified
as S1 or S2; the remaining identified frames are classified accordingly.

The nature of this algorithm requires that it work on short sequences, 4 to 6
seconds long, because as the sequence gets longer the periodicity of the
sequence fades away due to noise and variations of the heart rate.

To overcome this problem, the signal is split into 4-seconds wide windows and
the algorithm is applied to each window. The resulting sets of
heart sounds endpoint are then joined into a single set.

\begin{figure}[ht]
    \centering
    \includegraphics[width=0.5\columnwidth]{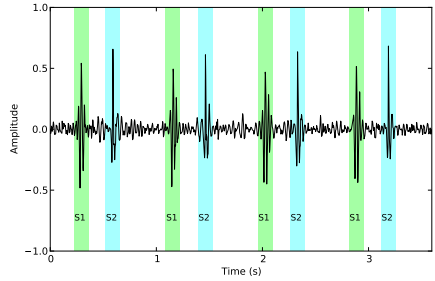}
    \caption{Example of S1 and S2 detection}
    \label{fig:segm}
\end{figure}

\subsubsection{The chirp $z$-transform}
\label{sec:czt}
The Chirp $z$-Transform (CZT) is an algorithm for the computation of the
$z$-Transform of sampled signals that offers some additional flexibility to
the Fast Fourier Transform (FFT) algorithm.

The main advantage of the CZT exploited in the analysis of heart sounds is
the fact that it allows high-resolution analysis of narrow frequency bands, offering
higher resolution than the FFT.

For more details on the CZT, please refer to \cite{rabiner-czt}

\subsubsection{Cepstral analysis}
\label{sec:cepstral}
Mel-Frequency Cepstrum Coefficients (MFCC) are one of the most
widespread parametric representation of audio signals (\cite{mfcc}).

The basic idea of MFCC is the extraction of cepstrum coefficients using a
non-linearly spaced filterbank; the filterbank is instead spaced according to
the Mel Scale: filters are linearly spaced up to 1 kHz, and then are
logarithmically spaced, decreasing detail as the frequency increases.

This scale is useful because it takes into account the way we perceive sounds.

The relation between the Mel frequency $\hat{f}_{mel}$ and the linear
frequency $f_{lin}$ is the following:

\begin{equation}
    \hat{f}_{mel} = 2595 \cdot \log_{10}\left(\frac{1 + f_{lin}}{700}\right)
\end{equation}

Some heart-sound biometry systems use MFCC, while others use a linearly-spaced
filterbank.

The first step of the algorithm is to compute the FFT of the input signal; the
spectrum is then feeded to the filterbank, and the $i$-th cepstrum coefficient
is computed using the following formula:

\begin{equation}
    \label{eq:mfcc}
    C_i = \sum_{k = 1}^{K} X_k \cdot \cos \left(i \cdot \left( k - \frac{1}{2}
\right) \cdot \frac{\pi}{K}\right) i = 0, ..., M
\end{equation}

where $K$ is the number of filters in the filterbank, $X_k$ is the
log-energy output of the $k$-th filter and $M$ is the number of coefficients
that must be computed.

Many parameters have to be chosen when computing cepstrum coefficients. Among
them: the bandwidth and the scale of the filterbank (Mel vs. linear), the
number and spectral width of filters, the number of coefficients.

In addition to this, differential cepstrum coefficients, tipically denoted
using a $\Delta$ (first order) or $\Delta\Delta$ (second order), can be
computed and used.

\begin{figure}[ht]
    \centering
    \includegraphics[width=.5\columnwidth]{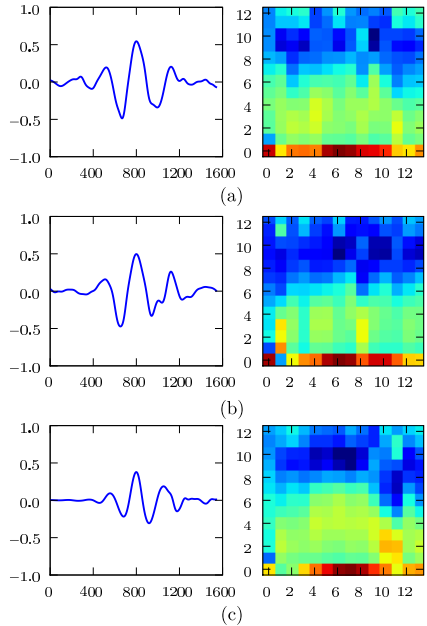}
    \caption{Example of waveforms and MFCC spectrograms of S1 sounds}
    \label{fig:mfcc_comp}
\end{figure}

Figure~\ref{fig:mfcc_comp} shows an example of three S1 sounds and the
relative MFCC spectrograms; the first two (a, b) belong to the same person,
while the third (c) belongs to a different person.

\subsubsection{The First-to-Second Ratio (FSR)}
\label{sec:fsr}
In addition to standard feature extraction techniques, it would be desirable to
develop ad-hoc features for the heart sound, as it is not a simple audio
sequence but has specific properties that could be exploited to develop
features with additional discriminative power.

This is why we propose a time-domain feature called First-to-Second Ratio
(FSR).  Intuitively, the FSR represents the power ratio of the first heart
sound (S1) to the second heart sound (S2). During our work, we observed that
some people tend to have an S1 sound that is louder than S2, while in others
this balance is inverted. We try to represent this diversity using our new
feature.

The implementation of the feature is different in the two biometric systems
that we described in this chapter, and a discussion of the two algorithms can
be found in \ref{sec:structural-fsr} and \ref{sec:statistical-fsr}.

\section{Review of related works}
\label{sec:review}
In the last years, different research groups have been studying the possibility of
using heart sounds for biometric recognition. In this section, we will briefly
describe their methods.

In Table~\ref{tab:papers-comparison} we summarized the main characteristics of
the works that will be analyzed in this section, using the following criteria:

\begin{itemize}
    \item{\textbf{Database} - the number of people involved in the study and the
    amount of heart sounds recorded from each of them;}
    \item{\textbf{Features} - which features were extracted from the signal, at frame
level or from the whole sequence;}
    \item{\textbf{Classification} - how features were used to make a decision.}
\end{itemize}

We chose not to represent performance in this table for two reasons: first,
most papers do not adopt the same performance metric, so it would be difficult
to compare them; second, the database and the approach used are quite
different one from another, so it would not be a fair comparison.

\begin{table}[h!]
  \begin{center}
    \begin{tabular}{l|l|c|c}
      \toprule
        \textbf{Paper} & \textbf{Database} & \textbf{Features} &  \textbf{Classification}\\
      \midrule
        \multirow{2}{*}{\cite{heart-singapore} } & 10 people   &  MFCC & GMM\\
                                                 & 100 HS each &  LBFC & VQ\\
      \midrule
        \multirow{2}{*}{\cite{tran:2010}}     & 52 people &  Multiple & SVM \\
                                              & 100m each &  & \\
      \midrule
        \multirow{2}{*}{\cite{jasper:2010}} & 10 people  &  Energy & Euclidean \\
                                            & 20 HS each &  peaks &  distance\\
      \midrule
        \multirow{3}{*}{\cite{fatemian:2010}} & 21 people  &  MFCC, LDA,    & Euclidean \\
                                              & 6 HS each &   energy peaks  & distance\\
                                              & 8 seconds per HS          &                 & \\
      \midrule
        \multirow{3}{*}{\cite{el-bendary:2010}} & 40 people    & autocorrelation & MSE \\
                                                & 10 HS &   cross-correlation & kNN\\
                                                & 10 seconds per HS          & complex cepstrum    & \\
      \bottomrule
    \end{tabular}
  \end{center}
  \caption{Comparison of recent works about heart-sound biometrics}
  \label{tab:papers-comparison}
\end{table}

In the rest of the section, we will briefly review each of these papers.

\cite{heart-singapore} was one of the first works in the field of heart-sounds
biometry. In this paper, the authors first do a quick exploration of the
feasibility of using heart sounds as a biometric trait, by recording a test
database composed of 128 people, using 1-minute heart sounds and splitting the same
signal into a train and a testing sequence. Having obtained good recognition
performance using the HTK Speech Recognition toolkit, they do a deeper test using
a database recorded from 10 people and containing 100 sounds for each person,
investigating the performance of the system using different feature extraction
algorithms (MFCC, Linear Frequency Band Cepstra (LFBC)), different classification schemes (Vector Quantization (VQ)
and Gaussian Mixture Models (GMM)) and investigating the impact of the
frame size and of the training/test length. After testing many combinations of
those parameters, they conclude that, on their database, the most performing
system is composed of LFBC features (60 cepstra + log-energy + 256ms frames with
no overlap), GMM-4 classification, 30s of training/test length.

The authors of \cite{tran:2010}, one of which worked on \cite{heart-singapore},
take the idea of finding a good and representative feature set for
heart sounds even further, exploring 7 sets of features: temporal shape, spectral shape, cepstral
coefficientrs, harmonic features, rhythmic features, cardiac features and the GMM
supervector. They then feed all those features to a feature selection method called
RFE-SVM and use two feature selection strategies (optimal and sub-optimal) to find
the best set of features among the ones they considered. The tests were conducted
on a database of 52 people and the results, expressed in terms of Equal Error Rate
(EER), are better for the automatically selected feature sets with respect to
the EERs computed over each individual feature set.

In \cite{jasper:2010}, the authors describe an experimental system where
the signal is first downsampled from 11025 Hz to 2205 Hz; then it is processed
using the Discrete Wavelet Transform, using the Daubechies-6 wavelet, and
the D4 and D5 subbands (34 to 138 Hz) are then selected for further processing. After
a normalization and framing step, the authors then extract from the signal some energy
parameters, and they find that, among the ones considered, the Shannon energy
envelogram is the feature that gives the best performance on their database of 10 people.

The authors of \cite{fatemian:2010} do not propose a pure-PCG approach, but they rather
investigate the usage of both the ECG and PCG for biometric recognition. In
this short
summary, we will focus only on the part of their work that is related to PCG. The
heart sounds are processed using the Daubechies-5 wavelet, up to the 5th scale, and
retaining only coefficients from the 3rd, 4th and 5th scales. They then use two
energy thresholds (low and high), to select which coefficients should be used
for further stages. The remaining frames are then processed using the
Short-Term Fourier Transform (STFT), the Mel-Frequency
filterbank and Linear Discriminant Analysis (LDA) for dimensionality reduction. The decision is made using
the Euclidean distance from the feature vector obtained in this way and the template
stored in the database. They test the PCG-based system on a database of 21 people,
and their combined PCG-ECG systems has better performance.

The authors of \cite{el-bendary:2010} filter the signal using the DWT; then
they extract different kinds of features: auto-correlation, cross-correlation
and cepstra. They then test the identities of people in their database, that is
composed by 40 people, using two classifiers: Mean Square Error (MSE) and k-Nearest Neighbor
(kNN). On their database, the kNN classifier performs better than the MSE one.

\section{The structural approach to heart-sounds biometry}
\label{sec:structural-approach}
The first system that we describe in depth was introduced in \cite{heart-bio};
it was designed to work with short heart sounds, 4 to 6 seconds long and thus
containing at least four cardiac cycles (S1-S2).

The restriction on the length of the heart sound was removed in \cite{heart-wifs09},
that introduced the quality-based best subsequence selection algorithm,
described in \ref{sec:structural-quality}.

We call this system ``structural'' because the identity templates are stored
as feature vectors, in opposition to the ``statistical'' approach, that
does not directly keep the feature vectors but instead it represents
identities via statistical parameters inferred in the learning phase.

\begin{figure*}[ht]
    \centering
    \scalebox{.7}{\input{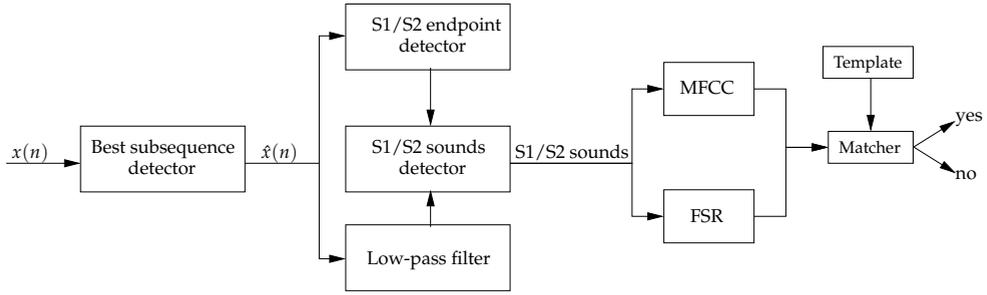}}
    \caption{Block diagram of the proposed cardiac biometry system}
    \label{fig:blockdiagram}
\end{figure*}

Figure~\ref{fig:blockdiagram} contains the block diagram of the system. Each
of the steps will be described in the following sections.

\subsection{The best subsequence selection algorithm}
\label{sec:structural-quality}
The fact that the segmentation and matching algorithms of the original system
were designed to work on short sequences was a strong constraint for the
system. It was required that a human operator selected a portion of the input
signal based on some subjective assumptions. It was clearly a flaw that needed
to be addressed in further versions of the system.

To resolve this issue, the authors developed a quality-based subsequence
selection algorithm, based on the definition of a quality index $DHS_{QI}(i)$
for each contiguous subsequence $i$ of the input signal.

The quality index is based on a cepstral similarity criterion: the selected subsequence
is the one for which the cepstral distance of the tones is the lowest
possible. So, for a given subsequence $i$, the quality index is defined as:

\begin{equation}
    DHS_{QI}(i) = \frac{1}{\displaystyle
        \sum_{k=1}^4 \sum_{\substack{j=1\\ j\neq k}}^4 d_{S1}(j, k) +
        \sum_{k=1}^4 \sum_{\substack{j=1\\ j\neq k}}^4 d_{S2}(j, k)}
\end{equation}

Where $d_{S1}$ and $d_{S2}$ are the cepstral distances defined in
\ref{sec:structural-matching}.

The subsequence $\overline{i}$ with the maximum value of
$DHS_{QI}(\overline{i})$ is then selected as the best one and retained for
further processing, while the rest of the input signal is discarded.

\subsection{Filtering and segmentation}
\label{sec:structural-segm}
After the best subsequence selection, the signal is then given in input to the
heart sound endpoint detection algorithm described in \ref{sec:segmentation}.

The endpoints that it finds are then used to extract the relevant portions of
the signal over a version of the heart sound signal that was previously filtered using a
low-pass filter, which removed the high-frequency extraneous components.

\subsection{Feature extraction}
The heart sounds are then passed to the feature extraction module, that
computes the cepstral features according to the algorithm described in
\ref{sec:heart-sounds-processing}.

This system uses $M = 12$ MFCC coefficients, with the addition of a 13-th
coefficient computed using an $i = 0$ value in Equation~\ref{eq:mfcc}, that is
the log-energy of the analyzed sound.

\subsection{Computation of the First-to-Second Ratio}
\label{sec:structural-fsr}
For each input signal, the system computes the FSR according to the following
algorithm.

Let $N$ be the number of complete S1-S2 cardiac cycles in the signal. Let
$P_{{S1}_i}$ (resp. $P_{{S2}_i}$) be the power of the $i$-th S1 (resp. S2)
sound.

We can then define $\overline{P_{S1}}$ and $\overline{P_{S2}}$, the average
powers of S1 and S2 heart sounds:

\begin{equation}
    \overline{P_{S1}} = \frac{1}{N} \sum_{i = 1}^{N} P_{S1_i}
\end{equation}
\begin{equation}
    \overline{P_{S2}} = \frac{1}{N} \sum_{i = 1}^{N} P_{S2_i}
\end{equation}

Using these definitions, we can then define the First-to-Second Ration of a
given heart sound signal as:

\begin{equation}
    FSR = \frac{\overline{P_{S1}}}{\overline{P_{S2}}}
\end{equation}

For two given DHS sequences $x_1$ and $x_2$, we define the FSR distance as:
\begin{equation}
    \label{eq:dfsr}
    d_{FSR}\left(x_1, x_2\right) = \left|FSR_{dB}\left(x_1\right) - FSR_{dB}\left(x_2\right)\right|
\end{equation}

\subsection{Matching and identity verification}
\label{sec:structural-matching}
The crucial point of identity verification is the computation of the distance
between the feature set that represents the input signal and the template
associated with the identity claimed in the acquisition phase by the person that
is trying to be authenticated by the system.

This system employs two kinds of distance: the first in the cepstral domain
and the second using the FSR.

MFCC are compared using the Euclidean metric ($d_2$). Given two heart sound
signals $X$ and $Y$, let $X_{S1}(i)$ (resp.  $X_{S2}(i)$) be the feature
vector for the $i$-th S1 (resp. S2) sound of the $X$ signal and $Y_{S1}$ and
$Y_{S2}$ the analogous vectors for the $Y$ signal. Then the cepstral distances
between $X$ and $Y$ can be defined as follows:

\begin{equation}
    d_{S1}(X, Y) = \frac{1}{N^2} \sum_{i,j = 1}^{N} d_2(X_{S1}(i), Y_{S1}(j))
\end{equation}
\begin{equation}
    d_{S2}(X, Y) = \frac{1}{N^2} \sum_{i,j = 1}^{N} d_2(X_{S2}(i), Y_{S2}(j))
\end{equation}

Now let us take into account the FSR. Starting from the $d_{FSR}$ as defined
in Equation~\ref{eq:dfsr}, we wanted this distance to act like an amplifying
factor for the cepstral distance, making the distance bigger when it has an
high value while not changing the distance for low values.

We then normalized the values of $d_{FSR}$ between 0 and 1
($d_{{FSR}_{norm}}$), we chose a threshold of activation of the FSR ($th_FSR$)
and we defined defined $k_{FSR}$, an amplifying factor used in the matching
phase, as follows:

\begin{equation}
    k_{FSR} = \max\left(1, \frac{d_{{FSR}_{norm}}}{th_{FSR}}\right)
\end{equation}

In this way, if the normalized FSR distance is lower than $th_{FSR}$ it has no
effect on the final score, but if it is larger, it will increase the cepstral
distance.

Finally, the distance between $X$ and $Y$ can be computed as follows:

\begin{equation}
    d(X, Y) = k_{FSR} \cdot \sqrt{d_{S1}(X, Y)^2 + d_{S2}(X, Y)^2}
\end{equation}

\section{The statistical approach to heart-sounds biometry}
\label{sec:statistical-approach}
In opposition to the system analyzed in Section~\ref{sec:structural-approach},
the one that will be described in this section is based on a learning
process that does not directly take advantage of the features extracted from
the heart sounds, but instead uses them to infer a statistical model of the
identity and makes a decision computing the probability that the input signal
belongs to the person whose identity was claimed in the identity verification
process.

\subsection{Gaussian Mixture Models}
\label{sec:statistical-gmm}
Gaussian Mixture Models (GMM) are a powerful statistical tool used for
the estimation of multidimensional probability density representation and
estimation (\cite{reynolds-gmm}).

A GMM $\lambda$ is a weighted sum of $N$ Gaussian probability densities:

\begin{equation}
    p(\bx | \lambda) = \sum_{i = 1}^{N} w_i p_i(\bx)
\end{equation}

where $\bx$ is a $D$-dimensional data vector, whose probability is being
estimated, and $w_i$ is the weight of the $i$-th probability density, that
is defined as:

\begin{equation*}
 p_i(\bx) = \frac{1}{\sqrt{(2\pi)^{D}\left| \Sigma_i \right|}} e ^ {- \frac{1}{2} (\bx - \mu_i)' \Sigma_i (\bx - \mu_i)}
\end{equation*}

The parameters of $p_i$ are $\mu_i$ ($\in \mathbb{R}^D$) and $\Sigma_i$ ($\in
\mathbb{R}^{D \times D}$), that together with $w_i$ ($\in \mathbb{R}^N$) form
the set of values that represent the GMM:

\begin{equation}
    \lambda = \left\{w_i, \mu_i, \Sigma_i\right\}
\end{equation}

Those parameters of the model are learned in the training phase using the
Expectation-Maximization algorithm (\cite{em}), using as input data the
feature vectors extracted from the heart sounds.

\subsection{The GMM/UBM method}
\label{sec:statistical-ubm}
The problem of verifying whether an input heart sound signal $s$ belongs to a
stated identity $I$ is equivalent to a hypothesis test between two hypotheses:
\begin{equation*}
    \begin{aligned}
        H_0: &\, s \mbox{  belongs to  } I\\
        H_1: &\, s \mbox{  does not belong to  } I\\
    \end{aligned}
\end{equation*}

This decision can be taken using a likelihood test:
\begin{equation}
\label{eq:decision}
    S(s,I) = \frac{p(s|H_0)}{p(s|H_1)} \left \{
    \begin{aligned}
        \geq \theta  & \mbox{  accept  } H_0\\
        < \theta  & \mbox{  reject  } H_0
    \end{aligned}
    \right.
\end{equation}

where $\theta$ is the decision threshold, a fundamental system parameter that
is chosen in the design phase.

The probability $p(s|H_0)$, in our system, computed using Gaussian Mixture Models.

The input signal is converted by the front-end algorithms to a set of $K$
feature vectors, each of dimension $D$, so:
\begin{equation}
    p(s|H_0) = \prod_{j = 1}^{K} p(x_j|\lambda_I)
\end{equation}

In Equation~\ref{eq:decision}, the $p(s|H_1)$ is still missing. In the
GMM/UBM framework (\cite{ubm}), this probability is modelled by building a
model trained with a set of identities that represent the demographic
variability of the people that might use the system. This model is called
Universal Background Model (UBM).

The UBM is created during the system design, and is subsequently used every
time the system must compute a matching score.

The final score of the identity verification process, expressed in terms of
log-likelihood ratio, is
\begin{equation}
    \label{eq:score}
    \Lambda(s) = \log S(s,I) = \log p(s|\lambda_I) - \log p(s|\lambda_W)
\end{equation}

\subsection{Front-end processing}
\label{sec:statistical-frontend}
Each time the system gets an input file, whether for training a model or for
identity verification, it goes through some common steps.

First, heart sounds segmentation is carried on, using the algorithm described
in Section~\ref{sec:segmentation}.

Then, cepstral features are extracted using a tool called \textit{sfbcep},
part of the SPro suite (\cite{nist-spro}). Finally, the FSR, computed as
described in Section~\ref{sec:statistical-fsr}, is appended to each feature
vector.

\subsection{Application of the First-to-Second Ratio}
\label{sec:statistical-fsr}
The FSR, as first defined in Section~\ref{sec:structural-fsr}, is a
sequence-wise feature, i.e., it is defined for the whole input signal. It is
then used in the matching phase to modify the resulting score.

In the context of the statistical approach, it seemed more appropriate to just
append the FSR to the feature vector computed from each frame in the feature
extraction phase, and then let the GMM algorithms generalize this knowledge.

To do this, we split the input heart sound signal in 5-second windows and we
compute an average FSR ($\overline{FSR}$) for each signal window. It is then
appended to each feature vector computed from frames inside the window.

\subsection{The experimental framework}
\label{sec:statistical-framework}
The experimental set-up created for the evaluation of this technique was
implemented using some tools provided by \alsp , an open source toolkit for
speaker recognition developed by the ELISA consortium between 2004 and 2008
(\cite{alize-spkdet}).

The adaptation of parts of a system designed for speaker recognition to a
different problem was possible because the toolkit is sufficiently
general and flexible, and because the features used for heart-sounds
biometry are similar to the ones used for speaker recognition, as outlined in
Section~\ref{sec:heart-sounds-processing}.

During the world training phase, the system estimates the parameters of the world
model $\lambda_W$ using a randomly selected subset of the input signals.

The identity models $\lambda_i$ are then derived from the world model $W$ using
the Maximum A-Posteriori (MAP) algorithm.

During identity verification, the matching score is computed using
Equation~\ref{eq:score}, and the final decision is taken comparing the score
to a threshold ($\theta$), as described in Equation~\ref{eq:decision}

\subsection{Optimization of the method}
\label{sec:statistical-optimization}
During the development of the system, some parameters have been tuned in order
to get the best performance. Namely, three different cepstral feature sets
have been considered in (\cite{heart-securware2010}):
\begin{itemize}
  \item 16 + 16 $\Delta$ + $E$ + $\Delta E$
  \item 16 + 16 $\Delta$ + 16 $\Delta \Delta$
  \item 19 + 19 $\Delta$ + $E$ + $\Delta E$
\end{itemize}
However, the first of these sets proved to be the most effective

In (\cite{heart-bioms2010}) the impact of the FSR and of the number of
Gaussian densities in the mixtures was studied. Four different model sizes
(128, 256, 512, 1024) were tested, with and without FSR, and the best
combination of those parameters, on our database, is 256 Gaussians with FSR.

\section{Performance evaluation}
\label{sec:comparison}
In this section, we will compare the performance of the two systems described
in Section~\ref{sec:structural-approach} and \ref{sec:statistical-approach}
using a common heart sounds database, that will be further described in
Section~\ref{sec:comparison-db}.

\subsection{Heart sounds database}
\label{sec:comparison-db}
One of the drawbacks of this biometric trait is the absence of large enough
heart sound databases, that are needed for the validation of biometric
systems. To overcome this problem, we are building a heart sounds database
suitable for identity verification performance evaluation.

Currently, there are 206 people in the database, 157 male and 49 female; for
each person, there are two separate recordings, each lasting from 20 to 70
seconds; the average length of the recordings is 45 seconds. The heart sounds
have been acquired using a Thinklabs Rhythm Digital Electronic Stethoscope,
connected to a computer via an audio card. The sounds have been converted
to the Wave audio format, using 16 bit per second and at a rate of 11025 Hz.

One of the two recordings available for each personused to build the
models, while the other is used for the computation of matching scores.

\subsection{Metrics for performance evaluation}
\label{sec:comparison-metrics}
A biometric identity verification system can be seen as a binary classifier.

Binary classification systems work by comparing matching scores to a
threshold; their accuracy is closely linked with the choice of the threshold,
which must be selected according to the context of the system.

There are two possible errors that a binary classifier can make:
\begin{itemize}
  \item \textbf{False Match (Type I Error):} accept an identity claim even if the
    template does not match with the model;
  \item \textbf{False Non-Match (Type II Error):} reject an identity claim even
    if the template matches with the model
\end{itemize}
The importance of errors depends on the context in which the biometric system
operates; for instance, in a high-security environment, a Type I error can be
critical, while Type II errors could be tolerated.

When evaluating the performance of a biometric system, however, we need to
take a threshold-independent approach, because we cannot know its
applications in advance. A common performance measure is the Equal Error Rate (EER)
(\cite{handbook-biometrics}), defined as the error rate at which the False
Match Rate (FMR) is equal to the False Non-Match Rate (FNMR).

A finer evaluation of biometric systems can be done by plotting the Detection
Error Tradeoff (DET) curve, that is the plot of FMR against FNMR. This allows
to study their performance when a low FNMR or FMR is imposed to the system.

The DET curve represents the trade-off between security and usability. A
system with low FMR is a highly secure one but will lead to more non-matches,
and can require the user to try the authentication step more times; a system
with low FNMR will be more tolerant and permissive, but will make more false
match errors, thus letting more unauthorized users to get a positive match.
The choice between the two setups, and between all the intermediate security
levels, is strictly application-dependent.

\subsection{Results}
\label{sec:comparison-results}
The performance of our two systems has been computed over the heart sounds
database, and the results are reported in Table~\ref{tab:performance-evaluation}.

\begin{table}[h!]
  \centering
    \begin{tabular}{l|l}
      \toprule
        \textbf{System} & \textbf{EER (\%)}\\
      \midrule
        Structural & 36.86\\
        Statistical & \textbf{13.66}\\
      \bottomrule
    \end{tabular}
  \caption{Performance evaluation of the two heart-sounds biometry systems}
  \label{tab:performance-evaluation}
\end{table}

The huge difference in the performance of the two systems reflects the fact
that the first one is not being actively developed since 2009, and it was
designed to work on small databases, while the second has already proved to
work well on larger databases.

It is important to highlight that, in spite of a 25\% increment of the size of
the database, the error rate remained almost constant with respect to the
last evaluation of the system, in which a test over a 165 people database
yielded a 13.70\% EER.

\begin{figure}[ht]
    \centering
    \includegraphics[width=0.5\columnwidth]{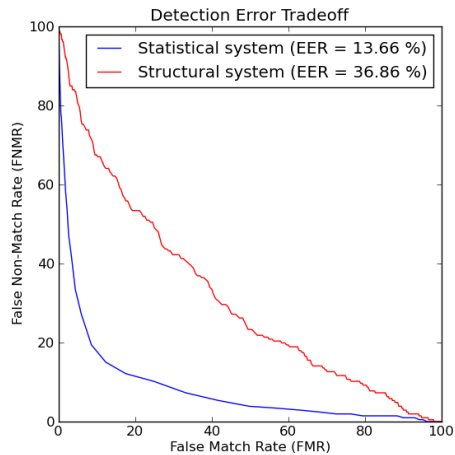}
    \caption{Detection Error Tradeoff (DET) curves of the two systems}
    \label{fig:det}
\end{figure}

Figure~\ref{fig:det} shows the Detection Error Trade-off (DET) curves of the
two systems. As stated before, a DET curve shows how the analyzed system
performs in terms of false matches/false non-matches as the system threshold
is changed.

In both cases, fixing a false match (resp. false non-match) rate, the system
that performs better is the one with the lowest false non-match (resp. false
match) rate.

Looking at Figure~\ref{fig:det}, it is easy to understand that the statistical
system performs better in both high-security (e.g., FMR = 1-2\%) and
low-security (e.g., FNMR = 1-2\%) setups.

We can therefore conclude that the statistical approach is definitely more
promising that the structural one, at least with the current algorithms and
using the database described in \ref{sec:comparison-db}..

\section{Conclusions}
\label{sec:conclusions}
In this chapter, we presented a novel biometric identification technique that
is based on heart sounds.

After introducing the advantages and shortcomings of this biometric trait with
respect to other traits, we explained how our body produces heart sounds, and
the algorithms used to process them.

A survey of recent works on this field written by other research groups has
been presented, showing that there has been a recent increase of interest
of the research community in this novel trait.

Then, we described the two systems that we built for biometric identification
based on heart sounds, one using a structural approach and another
leveraging Gaussian Mixture Models. We compared their performance over a
database containing more than 200 people, concluding that the statistical
system performs better.

\subsection{Future directions}
\label{sec:future-directions}
As this chapter has shown, heart sounds biometry is a promising research topic
in the field of novel biometric traits.

So far, the academic community has produced several works on this topic, but
most of them share the problem that the evaluation is carried on over small
databases, making the results obtained difficult to generalize.

We feel that the community should start a joint effort for the development of
systems and algorithms for heart-sounds biometry, at least creating a common
database to be used for the evaluation of different research systems over a
shared dataset that will make possible to compare their performance in order
to refine them and, over time, develop techniques that might be deployed in
real-world scenarios.

As larger databases of heart sounds become available to the scientific
community, there are some issues that need to be addressed in future
research.

First of all, the identification performance should be kept low even for larger
databases. This means that the matching algorithms will be fine-tuned and a
suitable feature set will be identified, probably containing both elements
from the frequency domain and the time domain.

Next, the mid-term and long-term reliability of heart sounds will be assessed,
analyzing how their biometric properties change as time goes by. Additionally,
the impact of cardiac diseases on the identification performance will be
assessed.

Finally, when the algorithms will be more mature and several independent
scientific evaluations will have given positive feedback on the idea, some
practical issues like computational efficiency will be tackled, and possibly
ad-hoc sensors with embedded matching algorithms will be developed, thus
making heart-sounds biometry a suitable alternative to the mainstream
biometric traits.

\end{document}